\documentclass[letterpaper, 10 pt, journal, twoside]{IEEEtran}

\IEEEoverridecommandlockouts                              

\usepackage{algorithm}
\usepackage{algorithmic}
\usepackage{amsmath} 
\usepackage{bm}
\usepackage{amssymb}
\usepackage[dvipdfmx]{color}
\usepackage{cite}
\usepackage[dvipdfmx]{graphicx} 
\usepackage{siunitx}

\newcommand{\figref}[1]{{Fig.~\ref{#1}}}
\newcommand{\tabref}[1]{{Tab.~\ref{#1}}}
\newcommand{\equref}[1]{{\eqref{#1}}} 

\newcommand{\refsec}[1]{Sec. \ref{sec:#1}}
\newcommand{\refsubsec}[1]{Sec. \ref{subsec:#1}}

\newcommand{\secref}[1]{\refsec{#1}} 
\newcommand{\subsecref}[1]{\refsubsec{#1}} 

\newif\ifdraft
\drafttrue 

\title{\LARGE \bf Design, Modeling and Control of a Quadruped Robot SPIDAR: Spherically Vectorable and Distributed Rotors Assisted \\ Air-Ground Amphibious Quadruped Robot}

\author{Moju Zhao$^{1}$, Tomoki Anzai$^{2}$, Takuzumi Nishio$^{2}$
  \thanks{$^{1}$ Department of Mechanical Engineering, The University of Tokyo, $^{2}$ Department of Mechano-Infomatics, The University of Tokyo, 7-3-1 Hongo, Bunkyo-ku, Tokyo 113-8656, Japan. {\tt\small chou@hnl.t.u-tokyo.ac.jp}}
  \thanks{Digital Object Identifier (DOI): see top of this page.}
}

\begin{document}

\maketitle

\markboth{}{}


\begin{abstract}
  Multimodal locomotion capability is an emerging topic in robotics field, and various novel mobile robots have been developed to enable the maneuvering in both terrestrial and aerial domains. Among these hybrid robots, several state-of-the-art bipedal robots enable the complex walking motion which is interlaced with flying.
  These robots are also desired to have the manipulation ability;
  however, it is difficult for the current forms to keep stability with the joint motion in midair due to the centralized rotor arrangement.
  Therefore, in this work, we develop a novel air-ground amphibious quadruped robot called SPIDAR which is assisted by spherically vectorable rotors distributed in each link to enable both walking motion and transformable flight.
  First, we present a unique mechanical design for quadruped robot that enables terrestrial and aerial locomotion.
  We then reveal the modeling method for this hybrid robot platform, and further develop an integrated control strategy for both walking and flying with joint motion.
  Finally, we demonstrate the feasibility of the proposed hybrid quadruped robot by performing a seamless motion that involves static walking and subsequent flight. To the best of our knowledge, this work is the first to achieve a quadruped robot with multimodal locomotion capability, which also shows the potential of manipulation in multiple domains.

\end{abstract}


\section{Introduction}
\label{sec:intro}

\begin{figure}[t]
  \begin{center}
    \includegraphics[width=1.0\columnwidth]{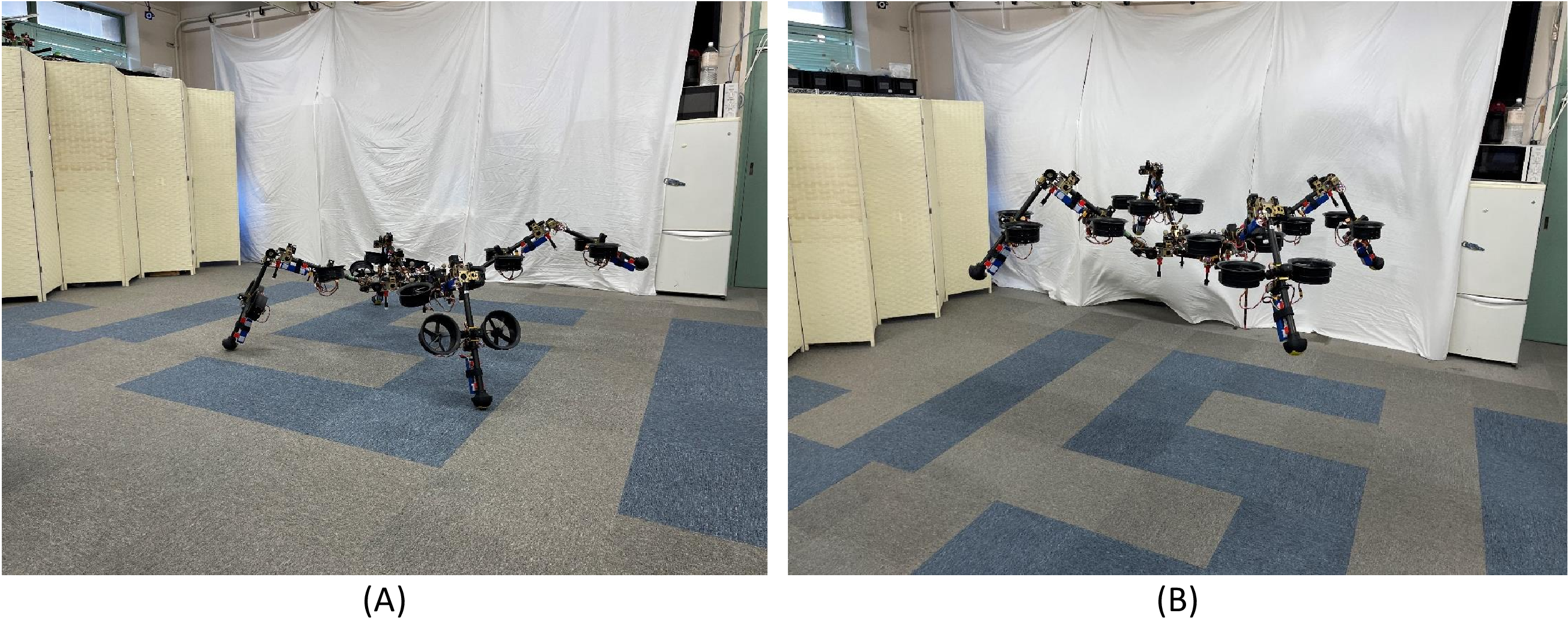}
    \vspace{-6mm}
    \caption{{\bf Air-ground amphibious quadruped robot SPIDAR}: {\bf SP}her{\bf I}cally vectorable and {\bf D}istributed rotors assisted {\bf A}ir-ground amphibious quadruped {\bf R}obot. {\bf (A)} walk on the ground like an usual quadruped robot, but with the assistance of thrust force; {\bf (B)} fly and transform in midair.}
    \label{figure:intro}
  \end{center}
\end{figure}

During the last decade, robots with multimodal locomotion capability have undergone intensive development, and demonstrated the versatile maneuvering in multiple domains (i.e., terrestrial, aerial, and aquatic) \cite{multimodal-locomotion:MUQA-IROS2013, multimodal-locomotion:qaudrotor-HyTAQ-IROS2014, multimodal-locomotion:Snake-ACM-R5-ISR2005,multimodal-locomotion:Salamander-Science2007, multimodal-locomotion:LoonCopter-JFR2018} which can benefit the performance in various situations, such as disaster response and industrial surveillance.
Among various forms for multimodal locomotion, the legged type has the advantage in unstructured terrains, and several bipedal models have been developed to demonstrate the promising walking that is interlaced with flying motion \cite{flying-bipedal-robot:LEONARDO-SciRo2021,flying-bipedal-robot:KXR-Humanoids2020}. The legged robots are desired to provide not only the advanced locomotion capability, but also the manipulation ability by the limb end-effector. Then, multilegged (multilimbed) is considered effective from the aspect of both the stability in the terrestrial locomotion and the freedom in manipulation.
In spite of the achievement of a simple arm motion in midair by the flying humanoid robot in \cite{flying-bipedal-robot:KXR-Humanoids2020}, the centralized rotor arrangement in most of the existing robot platforms can hardly handle the large change of Center-of-Gravity (CoG) caused by the joint motion in midair. Besides, the external force at the limb end during the aerial manipulation is also difficult to compensate by the centralized rotors due to the large moment arm.
Hence, a distributed rotor arrangement is necessary for aerial manipulation. 
Then, we propose a novel quadruped robot platform in this work, which can both walk and fly by using the spherically vectorable rotors distributed in each link unit as illustrated in \figref{figure:intro}.

One of the efficient type of air-ground hybrid robot platforms equips an ordinary multirotor for flying, and then deploys rolling cage or wheels to perform terrestrial locomotion
\cite{multimodal-locomotion:qaudrotor-HERALD-IROS2014, multimodal-locomotion:qaudrotor-HyTAQ-IROS2014, multimodal-locomotion:qaudrotor-FCSTAR-RAL2021}. Although the rolling mechanism can achieve the stable terrestrial locomotion without any complex control, it is relatively difficult for this type to handle the unstructured terrains (e.g., very few foothold). Then, legged model is proposed in several researches \cite{flying-bipedal-robot:LEONARDO-SciRo2021,flying-bipedal-robot:KXR-Humanoids2020, flying-bipedal-robot:Niiyama-ROBIO2018, flying-bipedal-robot:iCub-RAL2018, flying-bipedal-robot:Jet-HR2-RAL2022} to solve this issue by the walking motion. Most of these robots are bipedal model which attaches the multirotor unit in their torso for aerial maneuvering. Compared with the bipedal model, it is relatively easier to achieve the walking stability by the multilegged model because of the larger support polygon. Besides, the legs can be considered as the limbs for manipulation in midair, and thus more limbs can provide more end-effectors for complex manipulation task.
Therefore, we choose the quadruped model that can offer both the stable terrestrial locomotion and the potential of aerial manipulation.

For most of the quadruped robot, the form is bio-inspired and specialized for terrestrial locomotion \cite{quadruped-robot:LittleDog-IJRR2011, quadruped-robot:ANYmal-IROS2016, quadruped-robot:MiniCheetah-ICRA2019}. However, in our work, the robot is also desired to perform manipulation task in multiple domains. Several ape-like quadruped robots show the manipulation ability by the hands attached on the leg ends \cite{quadruped-robot:JPL-RoboSimian-JFR2015, quadruped-robot:WAREC-1-SSRR2017}.  The point symmetry is the crucial difference of these robots from the skeleton design for common quadruped robots. In our work, we also adopt the point symmetry for our hybrid quadruped robot to enable the omni-directional maneuvering and manipulation in both terrestrial and aerial domains.
Regarding the rotor arrangement, it is difficult for rotors centralized in the robot torso like \cite{flying-bipedal-robot:LEONARDO-SciRo2021} to handle the change of CoG caused by the joint motion. Besides, the external force acted at the end-effectors can also induce a large rotational load for the centralized rotors due to the large moment arm. To ensure a sufficient control margin for the stable joint motion in midair, \cite{flying-bipedal-robot:iCub-RAL2018} proposes a distributed rotor arrangement that deploys the thrust units at each limb end. However, this rotor arrangement deprives the robot of manipulation ability. Therefore, a fully-distributed rotor design proposed by \cite{aerial-robot:DRAGON-RAL2018} is applied in this work. In this design, spherically vectorable rotor apparatus is embedded in each link and thus can generate individual three-dimensional thrust force for the promising maneuvering and manipulation in midair as presented in \cite{aerial-robot:DRAGON-IJRR2022}.

For the stable flying motion, the whole platform should be lightweight, which leads a relatively compact and weak joint actuator for legs. Hence, the thrust force is also required to assist the walking motion by reducing the load from gravity. For model with centralized rotor arrangement, both the model-based \cite{flying-bipedal-robot:LEONARDO-SciRo2021} and the policy-based \cite{flying-bipedal-robot:KXR-IROS2022} methods are developed to obtain the assistive thrust. However, these methods are not suitable for the model with rotors distributed in all links. Then, \cite{flying-bipedal-robot:Jet-HR1-IROS2020} proposes an assistive thrust control method for a bipedal robot with the vectorable rotor attached at each foot, whereas \cite{aerial-robot:DRAGON-IJRR2022} presents a control method for a fully-distributed rotor model in the aerial domains.
Based on these methods, a comprehensive investigation of the modeling and control for the multilegged model is performed in this work to handle multiple types of torques and forces (i.e., the joint torque, the contact force on each limb end, the gravity of each link, and the thrust force from each vectorable rotor).


The main contributions of this work can be summarized as follows:
  \begin{enumerate}
  \item  We propose a unique mechanical design for air-ground quadruped robot with the spherically vectorable rotors distributed in all links.
  \item  We present a modeling and control methods for this multilegged platform for hybrid locomotion in both terrestrial and aerial domains.
  \item  We achieve the seamless and stable motion that involves walking, flying and joint motion in midair by the prototype of quadruped robot.
  \end{enumerate}


  The remainder of this paper is organized as follows. The mechanical design for this unique quadruped robot is introduced in \secref{design}. The modeling of our robot is presented in \secref{model}, followed by the integrated control method for hybrid locomotion in \secref{control}. We then show the experimental results in \secref{experiment} before concluding in \secref{conclusion}.

\section{Design}
\label{sec:design}

In this section, we present the mechanical design for the quadruped robot that is capable of terrestrial/aerial hybrid locomotion. The key of the whole structure is the spherically vectorable rotor embedded in each link unit, along with the unique quadruped shape that differs from the common bio-inspired type.

\begin{figure}[t]
  \begin{center}
    \includegraphics[width=1.0\columnwidth]{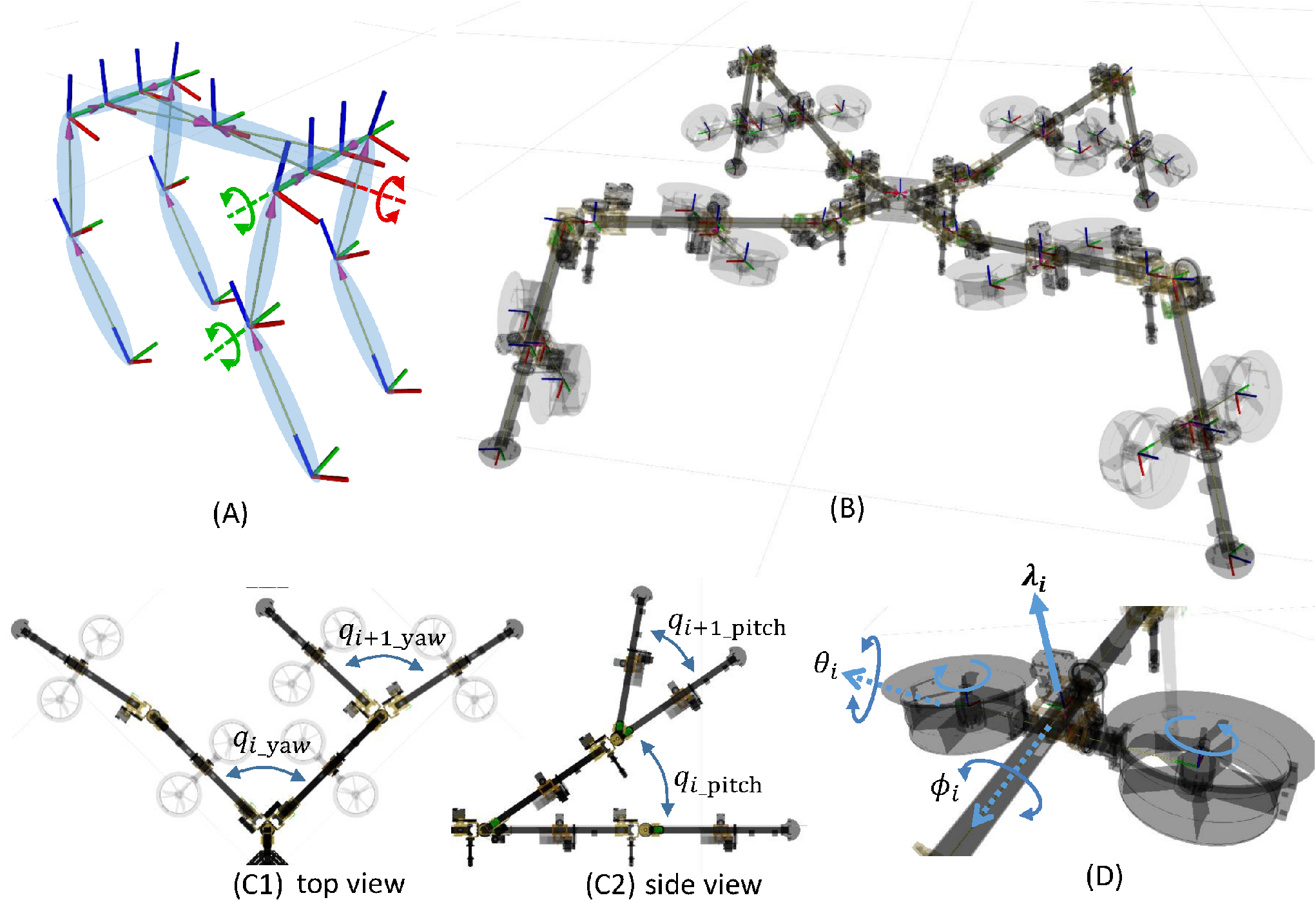}
    \caption{{\bf Mechanical design for the air-ground hybrid quadruped robot}. {\bf (A)} skeleton model for common bio-inspired mammal-type quadruped robot. {\bf (B)} proposed point-symmetric skeleton model for hybrid quadruped model. {\bf (C1)/(C2)} two-DoF joint module for each limb, where the yaw axis $q_{i\_\mathrm{yaw}}$ comes first followed by the pitch axis $q_{i\_\mathrm{pitch}}$. {\bf (D)} spherically vectorable rotor apparatus with two vectoring angles ($\phi, \theta$), and a combined thrust force $\lambda_i$ from the counter rotating dual rotors. There is a small offset between two vectoring axes because two rods cannot intersect with each other.}
    \label{figure:design}
  \end{center}
\end{figure}

\subsection{Skeleton Model}
\label{subsec:skeleton_model}

As shown in \figref{figure:design}(A), the common skeleton for quadruped robot is mammal-type, which puts a priority on the forward motion. Thus, the model is plane symmetric, and each leg has three DoF (two in the hip, and one in knee). However, our robot is desired to enable not only the terrestrial/aerial hybrid locomotion, but also the manipulation in midair. Therefore, the omni-directional movement is a critical feature for the skeleton design. Then, a point symmetric structure is introduced as depicted in \figref{figure:design}(B). This is similar to the sprawling-type quadruped design proposed by \cite{quadruped-robot:TITAN-XIII-IROS2013}, which can provide s wider supporting polygon and also a lower CoG than the mammal-type. According to this design concept, each limb consists of two links that have the same length, and is connected to the center torso with a joint module that has two Degree-of-Freedom (DoF). For this joint module, the yaw axis $q_{i\_\mathrm{yaw}}$ comes first, which is followed by the pitch axis $q_{i\_\mathrm{pitch}}$ to allow a larger swinging range for walking as shown in \figref{figure:design}(C1)/(C2). The crucial difference from the ordinary sprawling-type is that we also introduce an identical two-DoF joint module to connect neighboring links, which can provide a four-DoF manipulation capability by each limb end without the help of the torso motion. Eventually, this robot is composed of 8 links with 16 joints for walking and flying. Given that the lightweight design is significantly important for the flight performance, we deploy a compact servo motor to individually actuate each joint at the expense of the torque power. Nevertheless, the shortage of the joint torque can be compensated by the rotor thrust in our robot.

\subsection{Spherically Vectorable Rotor}
\label{subsec:vectorable_rotor}
Rotors embedded in links are used to achieve flight with arbitrary joint motion in midair. Besides, it is also necessary to use the rotor thrust to assist leg lifts for walking, because the joint actuator is weak due to the lightweight design as mentioned in \subsecref{skeleton_model}. Therefore, the rotor is required to point arbitrary direction to handle the change in link orientation. In other words, it is required to generate a three-dimensional thrust force by each rotor module to interact with not only the gravity and also the external force (i.e., the contact force on each foot). Then a spherically vectorable apparatus proposed by \cite{aerial-robot:DRAGON-RAL2018} is equipped in each link as depicted in \figref{figure:design}(D). 
To achieve the spherical vectoring around the link unit, two rotation axes is necessary. We first introduce a roll axis $\phi_i$ around the link rod. Then, we need the second orthogonal vectoring axis. If we use a single rotor, the collision between the propeller and the link rod will be inevitable while performing the second vectoring. Therefore, we apply the counter-rotating dual-rotor module to avoid the collision as shown in \figref{figure:design}(D). In addition, this dual-rotor can also counteract the drag moment and gyroscopic moment. Then, we define the pitch axis across dual rotors as the second vectoring axis $\theta_i$. Each vectoring axis is also actuated by an individual compact servo motor. Regarding the thrust force, since we assume that a pair of rotors rotate with the same speed, it is possible to introduced a combined thrust $\lambda_i$ for each spherically vectorable rotor module. Eventually, there are three control input (two vectoring angles $\phi_i, \theta_i$, and one thrust $\lambda_i$) for each vectorable rotor module, and totally 8 sets are used to control the whole quadruped model.

\section{Modeling}
\label{sec:model}

In this section, we describe the modeling for this robot which can be divided into two parts: the thrust model and the whole dynamics model. 

\begin{figure}[b]
  \begin{center}
    \includegraphics[width=1.0\columnwidth]{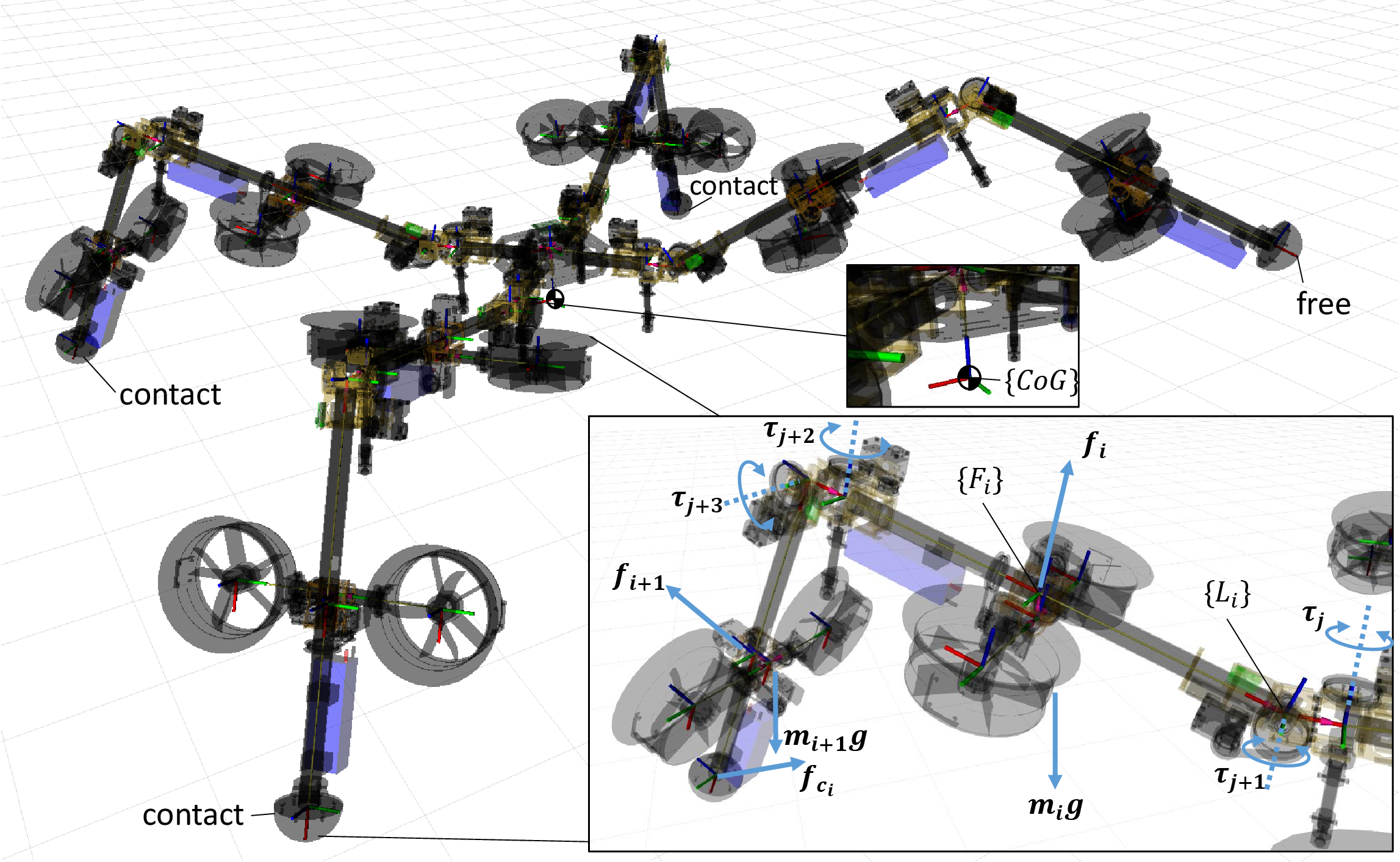}
    \caption{{\bf Dynamics model of the proposed quadruped robot}. The entire dynamics involves the joint torque ${\bm \tau}_j$, the contact force at each limb end ${\bm f}_{c_i}$, the gravity of each link $m_i\bm{g}$, and the thrust force from each vectorable rotor ${\bm f}_{i}$. $\{L_i\}$ and $\{F_i\}$ denote the frame for the $i$-th link and rotor respectively, whereas $\{CoG\}$ is the CoG frame for the whole model. For the free leg during walking, the contact force ${\bm f}_{c_i}$ at the limb end disappears.}
    \label{figure:model}
  \end{center}
\end{figure}

\subsection{Spherically Vectorable Thrust}
\label{subsec:allocation}
Based on the kinematic model depicted in \figref{figure:model}, the three-dimensional force $\bm{f}_i$ generated by the $i$-th rotor module can be written as:
\begin{align}
  \label{eq:thrust_model}
  \bm{f}_{i} &=  \lambda_i  \bm{u}_{i}, \\
  \label{eq:u_vector}
  \bm{u}_{i} &= {}^{CoG}R_{F_i}({\bm q}, \phi_i, \theta_i) \begin{bmatrix}0&0&1\end{bmatrix}^{\mathsf{T}},
\end{align}
where ${}^{CoG}R_{F_i}$ denotes a rotation matrix of the rotor frame $\{F_i\}$ w.r.t. the frame $\{CoG\}$. For this robot, we define the frame $\{CoG\}$ to have an origin at the CoG point as depicted in \figref{figure:model}, and a xyz coordinate that is identical to the baselink at the center torso.
$\bm{u}_{i}$ denotes the unit vector for the spherically vectorable mechanism that is effected by two vectoring angles $\phi_i$ and $\theta_i$.  Besides, this vector also depends on the joint angles ${\bm q} \in  {\mathcal R}^{N_J}$, $N_J$ is the number of joints.

Then the total wrench in the frame $\{CoG\}$ can be given by
\begin{align}
  \label{eq:total_wrench}
\begin{bmatrix} \bm{f}_{\lambda} \\ \bm{\tau}_{\lambda} \\ \end{bmatrix}
=
\begin{bmatrix}
 {\displaystyle \sum_{i = 1}^{N_r}} \bm{f}_{i} \\
 {\displaystyle \sum_{i = 1}^{N_r}} \bm{p}_{i} \times \bm{f}_{i} \\
\end{bmatrix}
= Q \bm{\lambda},
\end{align}
\begin{align}
  \label{eq:q_matrix}
Q &=
\begin{bmatrix}
  \bm{u}_1 & \bm{u}_2 & \cdots & \bm{u}_{N_{\mathrm{r}}} \\
  \bm{p}_{1} \times \bm{u}_1 & \bm{p}_{2} \times \bm{u}_2 & \cdots & \bm{p}_{N_{\mathrm{r}}} \times \bm{u}_{N_{\mathrm{r}}} \\
\end{bmatrix}
,  \\
 \bm{\lambda}  &=
\begin{bmatrix}
  \lambda_1 & \lambda_2 & \cdots & \lambda_{N_{\mathrm{r}}} \\
\end{bmatrix}^{\mathsf{T}}, \nonumber
\end{align}
where $\bm{p}_{i}$ is the position of the frame $\{F_i\}$ origin from the frame $\{CoG\}$ that is influenced by the joint angles $\bm{q}$ and the first vectoring angle $\phi_i$ for the $i$-th rotor. $N_{\mathrm{r}}$ is the number of rotors.

\subsection{Dynamics of Multilinked Model}
\label{subsec:dynamics}
The whole dynamic model can be written as follows:
\begin{align}
\label{eq:translational_dynamics}
 \dot{P}_{\Sigma} =& R_c \bm{f}_{\lambda} - m_{\Sigma} {\bm g} + {\displaystyle \sum_{i = 1}^{N_{\mathrm{c}}}} \bm{f}_{\mathrm{c}_i}, \\
\label{eq:rotational_dynamics}
 \dot{{\mathcal {\bm L}}}_{\Sigma} =&  \bm{\tau}_{\lambda} + {\displaystyle \sum_{i = 1}^{N_{\mathrm{c}}}} \bm{p}_{\mathrm{c}_i} \times R_c^{\mathsf{T}} \bm{f}_{\mathrm{c}_i}, \\
\label{eq:joint_dynamics}
M_J(\bm{q})\ddot{\bm{q}} + c(\bm{q},\dot{\bm{q}})  =& \displaystyle \bm{\tau}_{q}
  + \sum^{N_{\mathrm{c}}}_{i=1} J_{\mathrm{c}_i}^{\mathsf{T}} \bm{f}_{\mathrm{c}_i} \nonumber  \\
& +  \sum^{N_r}_{i=1} J_{\mathrm{r}_i}^{\mathsf{T}} \bm{f}_i
  + \sum^{N_s}_{i=1} J_{\mathrm{s}_i}^{\mathsf{T}} m_{\mathrm{s}_i}\bm{g}.
\end{align}

\equref{eq:translational_dynamics} and \equref{eq:rotational_dynamics} denote the centroidal dynamics for the whole multibody model.
$P_{\Sigma}$ and ${\mathcal {\bm L}}_{\Sigma}$ are the  entire linear and  rotational momentum described in the inertial frame $\{W\}$ and the frame $\{CoG\}$, respectively.
These momentum are both affected by the joint angles, vectoring angles, and their velocities (i.e., $\bm{q}, \dot{\bm{q}}, \bm{\phi}, \dot{\bm{\phi}}, \bm{\theta}, \dot{\bm{\theta}}$).
$R_c$ is the orientatin of the frame $\{CoG\}$ w.r.t. the frame $\{W\}$, and is identical to $R_b$ that is the orientatin of the baselink. 
${\bm f}_{\lambda}$ and ${\bm \tau}_{\lambda}$ corresponds to the total wrench described in \equref{eq:total_wrench}.
$\bm{f}_{\mathrm{c}_i}$ is the contact force at the $i$-th limb end (foot) w.r.t. the frame $\{W\}$, whereas $\bm{p}_{\mathrm{c}_i}$ is the position of this contact point from the frame $\{CoG\}$ which is also influenced by the joint angles $\bm{q}$. $N_{\mathrm{c}}$ is the number of contact points (i.e., standing legs).
$m_{\Sigma}$ is the total mass, and $\bm{g}$ is a three-dimensional vector expressing gravity.

\equref{eq:joint_dynamics} corresponds to the joint motion. $M_J(\bm{q})$ denotes the inertial matrix, whereas $c(\bm{q},\dot{\bm{q}})$ is the term related to the centrifugal and Coriolis forces in joint motion.
$J_{{\mathrm{*}}_i}\in{\mathcal R}^{3 \times N_J}$ is the Jacobian matrix for the frame of the $i$-th contact point ($* \rightarrow \mathrm{c}$), the $i$-th rotor ($* \rightarrow \mathrm{r}$), and the $i$-th segment's CoG ($* \rightarrow \mathrm{s}$), respectively. $\bm{\tau}_q \in \mathcal{R}^{N_J}$ is the vector of joint torque and $\bm{f}_i$ is the vectoring thrust force corresponding to \equref{eq:thrust_model}.

\eqref{eq:translational_dynamics} $\sim$ \equref{eq:joint_dynamics} are highly complex due to the joint motion. Then the realtime feedback control based on these nonlinear equations is significantly difficult for an onboard computational resource. Therefore, for the joint motion, a crucial assumption is introduced in our work to simplify the dynamics, i.e., all the joints are actuated slowly by individual servo motors. This is also called the quasi-static assumption that allows $\dot{\bm{q}} \approx \bm{0}; \ddot{\bm{q}} \approx \bm{0}$ during the joint motion.

Under this assumption, the original dynamic model can be approximated as follows:
\begin{align}
  \label{eq:approx_translational_dynamics}
&m_{\Sigma} \ddot{\bm r}_c(\bm{q}) = R_c \bm{f}_{\lambda}
- m_{\Sigma} {\bm g} + {\displaystyle \sum_{i = 1}^{N_{\mathrm{c}}}} \bm{f}_{\mathrm{c}_i}, \\
\label{eq:approx_rotational_dynamics}
&I_{\Sigma}(\bm{q})\dot{\bm \omega} + {\bm \omega} \times I_{\Sigma}(\bm{q}) {\bm \omega} = \bm{\tau}_{\lambda} + {\displaystyle \sum_{i = 1}^{N_{\mathrm{c}}}} \bm{p}_{\mathrm{c}_i} \times \bm{f}_{\mathrm{c}_i}, \\
\label{eq:approx_joint_dynamics}
& \bm{0} = \displaystyle \bm{\tau}_{q}
  + \sum^{N_{\mathrm{c}}}_{i=1} J_{\mathrm{c}_i}^{\mathsf{T}} \bm{f}_{\mathrm{c}_i}
  +  \sum^{N_r}_{i=1} J_{\mathrm{r}_i}^{\mathsf{T}} \bm{f}_i
  + \sum^{N_s}_{i=1} J_{\mathrm{s}_i}^{\mathsf{T}} m_{\mathrm{s}_i}\bm{g},
\end{align}
where $\bm{r}_c$ is the position of the frame $\{CoG\}$ w.r.t the frame $\{W\}$, which can be calculated using the forward-kinematics from the baselink states with the joint angles $\bm{q}$. ${\bm \omega}_c$ is the angular velocity of the frame $\{CoG\}$  w.r.t the frame of $\{CoG\}$, and is identical to ${\bm \omega}_b$ that is the angular velocity of the baselink. $I_{\Sigma} (\bm{q})$ is the total inertia tensor that is also influenced by the joint angles $\bm{q}$.

\equref{eq:approx_translational_dynamics} and \eqref{eq:approx_rotational_dynamics} show the property of a single rigid body, whereas \equref{eq:approx_joint_dynamics} indicates the equilibrium between the forces and torques for the joint motion. Given that we apply the quasi-static assumption for joint motion, only the slow terrestrial motion, such as the static walk gait, is allowed.

\section{Control}
\label{sec:control}

\begin{figure}[b]
  \begin{center}
    \includegraphics[width=1.0\columnwidth]{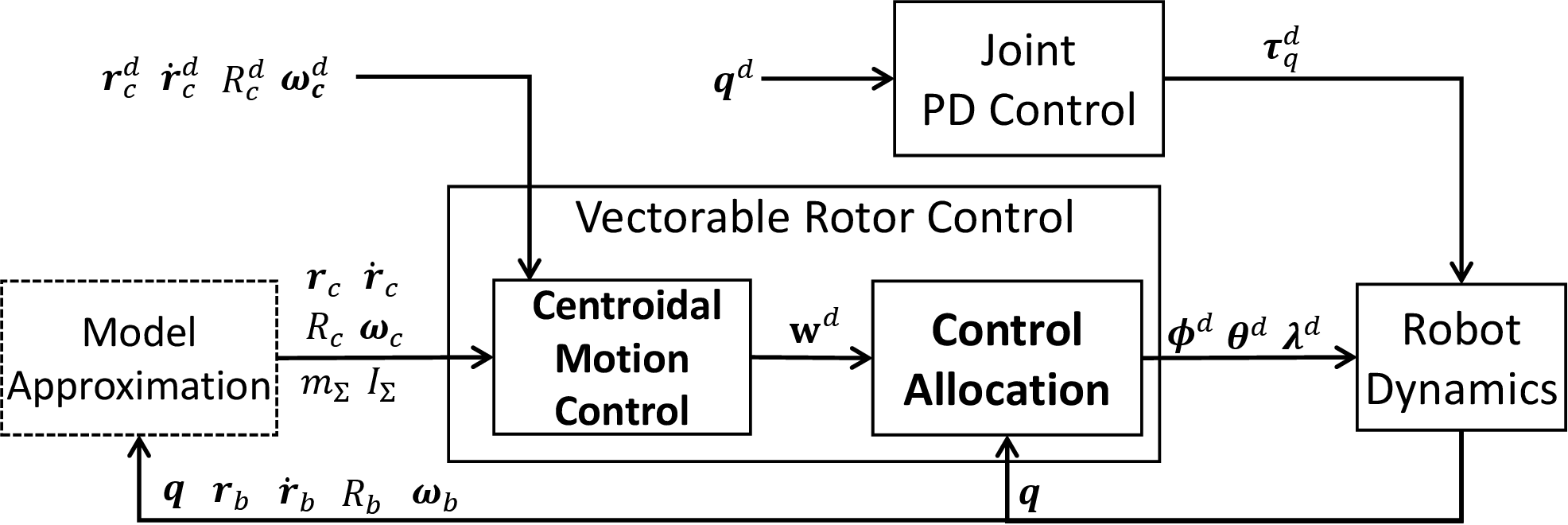}
    \caption{{\bf Unified control framework for terrestrial/aerial locomotion}. ``Model approximation'' presented in \subsecref{dynamics} is followed by the vectorable rotor control based on the centroidal motion. The joint control is performed independently.}
    \label{figure:control}
  \end{center}
\end{figure}

In this section, we first describe a unified control framework as depicted in \figref{figure:control}, and then present the modification for the aerial and terrestrial locomotion, respectively.

\subsection{Centroidal Motion Control}

For the approximated dynamics of \eqref{eq:approx_translational_dynamics} and \eqref{eq:approx_rotational_dynamics}, the position feedback control based on an ordinary PID control is given by
\begin{align}
  \label{eq:pid_pos}
  {\bm f}_{\lambda}^d &=  m_{\Sigma} R_c^{\mathsf{T}} (K_{f, p} \bm{e}_{\bm{r}} + K_{f, i} \int \bm{e}_{\bm{r}} + K_{f, d} \dot{\bm{e}}_{\bm{r}}) \nonumber \\
  & \hspace{5mm} + R_c^{\mathsf{T}} (m_{\Sigma} {\bm g} - {\displaystyle \sum_{i = 1}^{N_{\mathrm{c}}}} \bm{f}_{\mathrm{c}_i}),
\end{align}
where $\bm{e}_{\bm{r}} = \bm{r}_c^d - \bm{r}_c$, and $K_{f, \ast}$ are the PID gain diagonal matrices.

The attitude control follows the SO(3) control method proposed by \cite{aerial-robot:SE3-Control-CDC2010}:
\begin{align}
  \label{eq:pid_rot}
  {\bm \tau}^d_{\lambda} &= I_{\Sigma} (K_{\tau, p} \bm{e}_{R} + K_{\tau, i} \int \bm{e}_{R} + K_{\tau, d} \bm{e}_{\bm{\omega}}) \nonumber \\
  & \hspace{3mm}+ {\bm \omega}_c \times I_{\Sigma} {\bm \omega}_c -{\displaystyle \sum_{i = 1}^{N_{\mathrm{c}}}}  \bm{p}_{\mathrm{c}_i} \times R_c^{\mathsf{T}} \bm{f}_{\mathrm{c}_i}, \\
  \label{eq:rotation_error}
  \bm{e}_{R} &= \frac{1}{2}\left[R_c^{\mathsf{T}}R_c^d - R_c^{d\mathsf{T}}R_c\right]^{\vee}, \\
  \label{eq:omega_error}
  \bm{e}_{\bm{\omega}} &= R_c^{\mathsf{T}}R_c^d\bm{\omega}_c^d - \bm{\omega}_c,
\end{align}
where $\left[\star\right]^{\vee}$ is the inverse of a skew map.

Then, the desired wrench w.r.t the frame $\{CoG\}$ can be summarized as follows:
\begin{align}
  \label{eq:desired_wrench}
  \bm{\mathrm{w}} ^d = \begin{bmatrix} \bm{f}_{\lambda} ^d && \bm{\tau}_{\lambda} ^d \end{bmatrix}^{\mathsf{T}}.
\end{align}

\subsection{Control Allocation}

The goal of vectorable rotor control is to obtain the control input (the desired thrust $\bm{\lambda}^d$ and the desired vectoring angles $\bm{\phi}^d, \bm{\theta}^d$) from the desired wrench ${\bm{\mathrm{w}}}^d$. Meanwhile, it is also important to suppress the rotor output and the joint load from the aspect of the energy consumption. Then, an optimization problem should be design to obtain the desired control input. 
Since the vectoring angles $\bm{\phi}$ and $\bm{\theta}$ demand the trigonometric function, nonlinear constraints would appear in the optimization problem and thus lead a complex computation. To decrease the computational load during the realtime control loop, we introduce an alternative three-dimensional forces $\bm{f}^{'}_{i}$ that is the vectorable thrust w.r.t, the related link frame $\{L_i\}$: $\bm{f}^{'}_{i} = {}^{L_i}R_{F_i}(\phi_i, \theta_i) \begin{bmatrix}0&0&\lambda_i\end{bmatrix}^{\mathsf{T}}$. The definition of the frames of $\{L_i\}$ and $\{F_i\}$ can be found in \figref{figure:model}. Then the above optimization problem can be modified as follows:
\begin{align}
  \label{eq:rough_allocation_cost}
&   \displaystyle \min_{\bm{f}^{'}_{i}, \bm{\tau}_q, \bm{f}_{c_i}}  \hspace{3mm} w_1  {\displaystyle \sum_{i = 1}^{N_{\mathrm{r}}}} \| \bm{f}^{'}_{i} \|^2 + w_2 \|\bm{\tau}_q\|^2,\\
  \label{eq:wrench_allocation_constraint}
&   s.t.   \hspace{3mm}
  \bm{\mathrm{w}} ^d  = {\displaystyle \sum_{i = 1}^{N_{\mathrm{r}}}} {Q}_{i} \bm{f}^{'}_{i},
  \hspace{3mm}
  {Q}_{i} =
  \begin{bmatrix}
    E_{3\times 3} \\
    \left[\bm{p}_i \times \right]
  \end{bmatrix} {}^{CoG}R_{L_i},\\
  \label{eq:quasi_static_constraint}
  &  \hspace{8mm} \displaystyle \bm{\tau}_{q} =
  - \sum^{N_{\mathrm{c}}}_{i=1} J_{\mathrm{c}_i}^{\mathsf{T}} \bm{f}_{\mathrm{c}_i}
  -  \sum^{N_{\mathrm{r}}}_{i=1} J_{\mathrm{r}_i}^{\mathsf{T}} \bm{f}_i
  - \sum^{N_s}_{i=1} J_{\mathrm{s}_i}^{\mathsf{T}} m_{\mathrm{s}_i}\bm{g}, \\
  \label{eq:thrust_constraint}
  & \hspace{8mm} 0 < \lambda_i < \bar{\lambda},  \\
  \label{eq:joint_constraint}
  & \hspace{8mm} - \bar{\tau}_q < \tau_{q_i} < \bar{\tau}_q, \\
  \label{eq:contact_force_constraint}
  & \hspace{8mm} 0 < f_{c_i}(2),
\end{align}
where $w_1$ and $w_2$  in \equref{eq:rough_allocation_cost} are the weights for the cost of rotor thrust and joint torque, respectively.
\equref{eq:wrench_allocation_constraint} is the modified form of wrench allocation from \equref{eq:total_wrench} by using the alternative variable $\bm{f}^{'}_{i}$. $\bm{p}_i$ is defined in \equref{eq:total_wrench}, whereas ${}^{CoG}R_{L_i}$ is the orientation of the frame $\{L_i\}$ w.r.t. the frame $\{CoG\}$. $E_{3\times 3}$ is a 3 $\times$ 3 identity matrix and $\left[\cdot \times \right]$ denotes the skew symmetric matrix of a three dimensional vector.
\equref{eq:quasi_static_constraint} denotes the equilibrium between the joint torque $\bm{\tau}_q$, the contact force $\bm{f}_{c_i}$, the thrust force $\bm{f}_i$, and the segment gravity $m_{\mathrm{s}_i}\bm{g}$ to satisfy the joint quasi-static assumption. \equref{eq:thrust_constraint} and \equref{eq:joint_constraint} denote the bounds for the rotor thrust and joint torque, respectively.  The contact force $\bm{f}_{c_i}$ is also considered as the searching variable, and the $z$ element $f_{c_i}(2)$ should be always non-negative as shown in \equref{eq:contact_force_constraint}.

Given that all constraints \equref{eq:wrench_allocation_constraint} $\sim$ \equref{eq:contact_force_constraint} are linear, an ordinary algorithm for quadratic problem can be applied. Once the optimized thrust force $\tilde{\bm{f}}^{'}_{i}$ is calculated, the true control input for the spherically vector rotor apparatus can be obtained as follows:
\begin{align}
  \label{eq:pseudo_inverse_desired_thrust_force}
  &\lambda_i = \| {\bm{f}}^{'}_{i} \| ,\\
  \label{eq:pseudo_inverse_desired_vectoring_phy}
  &\phi_i = tan^{-1}(\frac{-f^{'}_{i}(1)}{f^{'}_{i}(2)}),\\
  \label{eq:pseudo_inverse_desired_vectoring_theta}
  &\theta_{i} = tan^{-1}(\frac{f^{'}_{i}(0)}{-f^{'}_{i}(1) sin(\phi_{i}) + f^{'}_{i}(2) cos(\phi_{i})}),
\end{align}
where $f^{'}_{i}(0)$, $f^{'}_{i}(1)$, and $f^{'}_{i}(2)$ are the $x,y$, and $z$ element of the vector.

As a unique mechanical feature of the spherically vectorable apparatus depicted in \figref{figure:design}(D), the result of vectoring angles $\bm{\phi}$ and  $\bm{\theta}$ from \eqref{eq:pseudo_inverse_desired_vectoring_phy} and \eqref{eq:pseudo_inverse_desired_vectoring_theta} will deviate the position $\bm{p}_i$ in \eqref{eq:wrench_allocation_constraint} because of the small offset between two vectoring axes as depicted in \figref{figure:design}(D). Then, the results of \eqref{eq:pseudo_inverse_desired_thrust_force} $\sim$ \eqref{eq:pseudo_inverse_desired_vectoring_theta} will no longer satisfy the constraint  \eqref{eq:wrench_allocation_constraint} because ${Q_i}$ has changed. To solve this problem, we apply the iteration process that is based on the gradient of a residual term  $\bm{\epsilon} := \bm{\mathrm{\mathrm{w}}} ^d - Q(\bm{\theta}, \bm{\phi}) \bm{\lambda}$, and finally we can obtain the convergent values of $\bm{\phi}^d$, $\bm{\theta}^d$ and $\bm{\lambda}^d$. The detail can be found in \cite{aerial-robot:DRAGON-IJRR2022}.

\subsection{Joint Control}
The proposed optimization problem of \equref{eq:rough_allocation_cost} can also provides the joint torque that however only satisfies the quasi-static assumption for joint motion. In addition, the measurement bias and noisy from the joint encoders along with the slight deformation of the link and joint structure can also induce the model error. To handle this model error, it is necessary to apply a feed-back control to track the desired position for joints. Therefore, a simple PD control for joint position is introduced for each joint:
\begin{align}
  \label{eq:joint_pd_control}
\tau_i^d  =  k_{\mathrm{j},p}(q_i^d - q_i) - k_{\mathrm{j},d} \dot{q}_i,
\end{align}
where $q_i^d$ is the desired joint angle from the walking gait or the aerial transformation planning. $k_{\mathrm{j},p}$ and $k_{\mathrm{j},d}$ are the P and D control gains. 
It is also notable that we also used the same PD control for the rotor vectoring angles $\phi_i$ and $\theta_i$.

\subsection{Aerial Locomotion}
\label{subsec:flight_control}

The control mode for aerial locomotion follows the flow shown in \figref{figure:control} but without the contact force $\bm{f}_{c_i}$. Then the constraint of \equref{eq:contact_force_constraint} and the first term $\sum^{N_{\mathrm{c}}}_{i=1} J_{\mathrm{c}_i}^{\mathsf{T}} \bm{f}_{\mathrm{c}_i}$ at the right side of \equref{eq:quasi_static_constraint} can be omitted. The joint control is executed independently to follow the trajectory given by other task planing.

\subsection{Terrestrial Locomotion}
\label{subsec:walk_control}

\subsubsection{Torso altitude control}

The terrestrial locomotion is totally based on the quasi-static joint motion. Therefore the centroidal motion should be also assumed to be static, which results in a desired wrench only handling gravity ($\bm{\mathrm{w}} ^d =  \begin{bmatrix} 0 & 0 & -m_{\Sigma}\bm{g}  & 0 & 0 & 0 \end{bmatrix}$) for \equref{eq:wrench_allocation_constraint}.
Despite of the joint position control proposed in \equref{eq:joint_pd_control}, a small error regarding the torso (i.e., baselink) pose, particularly along the altitude direction, would still remain mainly due to the influence of gravity. Therefore, we apply a feedback control using the rotor thrust for the torso altitude. Instead of the PID position control for the centroidal motion as proposed in \equref{eq:pid_pos}, a truncated feedback control for the torso altitude is introduced as follows:
\begin{align}
  \label{eq:pid_torso_altitude}
  f^d_z = k_{b} (z_b^d - z_b),
\end{align}
where $k_{b}$ is the P gain, and $z_b$ is the torso altitude.
We assume this altitude control is for the ``floating'' baselink even in the terrestrial locomotion mode. Therefore, instead of considering $f^d_z$ in \equref{eq:wrench_allocation_constraint} and \equref{eq:quasi_static_constraint}, we introduce another independent control allocation to obtain the additional thrust force as follows:
\begin{align}
  \label{eq:torso_altitude_allocation}
  \Delta \bm{\mathrm{w}} ^d = {\displaystyle \sum_{i = 1}^{\frac{N_{\mathrm{r}}}{2}}} {Q}_{2i} \Delta \bm{f}^{'}_{2i},
\end{align}
where $\Delta \bm{\mathrm{w}} ^d =  \begin{bmatrix} 0 & 0 & f^d_z & 0 & 0 & 0 \end{bmatrix}^{\mathsf{T}}$. It is notable that we only choose the rotors in the inner link of each leg to suppress the influence on the joint quasi-static motion as presented in \equref{eq:quasi_static_constraint}. Then $\Delta \bm{f}^{'}_{2i}$ can be given by
\begin{align}
  \label{eq:torso_altitude_allocation_inv}
  \Delta \bm{f}^{'} &= \tilde{Q}^{\#} \Delta \bm{\mathrm{w}} ^d, \\
  \tilde{Q} &= \begin{bmatrix} {Q}_{0} & {Q}_{2} & \cdots & {Q}_{N_{\mathrm{r}}} \end{bmatrix}, \nonumber \\
  \Delta \bm{f}^{'} &= \begin{bmatrix} \Delta \bm{f}^{'}_{0} & \Delta \bm{f}^{'}_{2} & \cdots & \Delta \bm{f}^{'}_{N_{\mathrm{r}}} \end{bmatrix}^{\mathsf{T}}, \nonumber 
\end{align}
where $\tilde{Q}^{\#}$ is the psuedo-inverse matrix of $\tilde{Q}$. Finally, $\bm{f}^{'}_{2i}  \rightarrow \bm{f}^{'}_{2i} + \Delta \bm{f}^{'}_{2i}$ is performed before substituting it into \equref{eq:pseudo_inverse_desired_thrust_force}$\sim$\equref{eq:pseudo_inverse_desired_vectoring_theta}.

\subsubsection{Static walking gait}
In this work, we only focus on the static walking gait. Hence only one leg is allowed to lift during walking. As the update of the foot step for the lifting leg, we analytically solve the inverse-kinematics for the related three joint angles: $q_{i\_\mathrm{yaw}}, q_{i\_\mathrm{pitch}}$, and $q_{i+1\_\mathrm{pitch}}$ as depicted in \figref{figure:design}, which can be uniquely determined. Regarding the gait for linear movement, we design a creeping gait that lifts the front-left, front-right, rear-right, and rear-left legs in order for one gait cycle, and also solely moves the torso in standing mode just after the two front legs have moved to the new position. To enable the repetition of the gait cycle, the stride length of all feet is set equal to the moving distance of torso.

We further assume the robot only walks on a flat floor, and thus the height of feet should be always zero. Then, we first set an intermediate target position right above the new foot step with a small height offset. Thus $q_{i\_\mathrm{yaw}}$ and $q_{i+1\_\mathrm{pitch}}$ are identical to the final target, whereas $q_{i\_\mathrm{pitch}}$ is smaller than the final value. Once the lifting leg moves to this intermediate pose, the robot starts lowering the leg to reach the new foot step only by changing $q_{i\_\mathrm{pitch}}$. Given that there is no tactile sensor on the foot, we introduce a threshold $\Delta q_{c}$ for the joint angle error of $q_{i\_\mathrm{pitch}}$ to detect touchdown. That is, if $q^d_{i\_\mathrm{pitch}} - q_{i\_\mathrm{pitch}} < \Delta q_{c}$, then switch the lifting leg to the standing mode, and thus the number of the contact force $\bm{f}_{c_i}$ changes from three to four.

\section{Experiment}
\label{sec:experiment}

\subsection{Robot Platform}

\begin{figure}[b]
  \begin{center}
    \includegraphics[width=1.0\columnwidth]{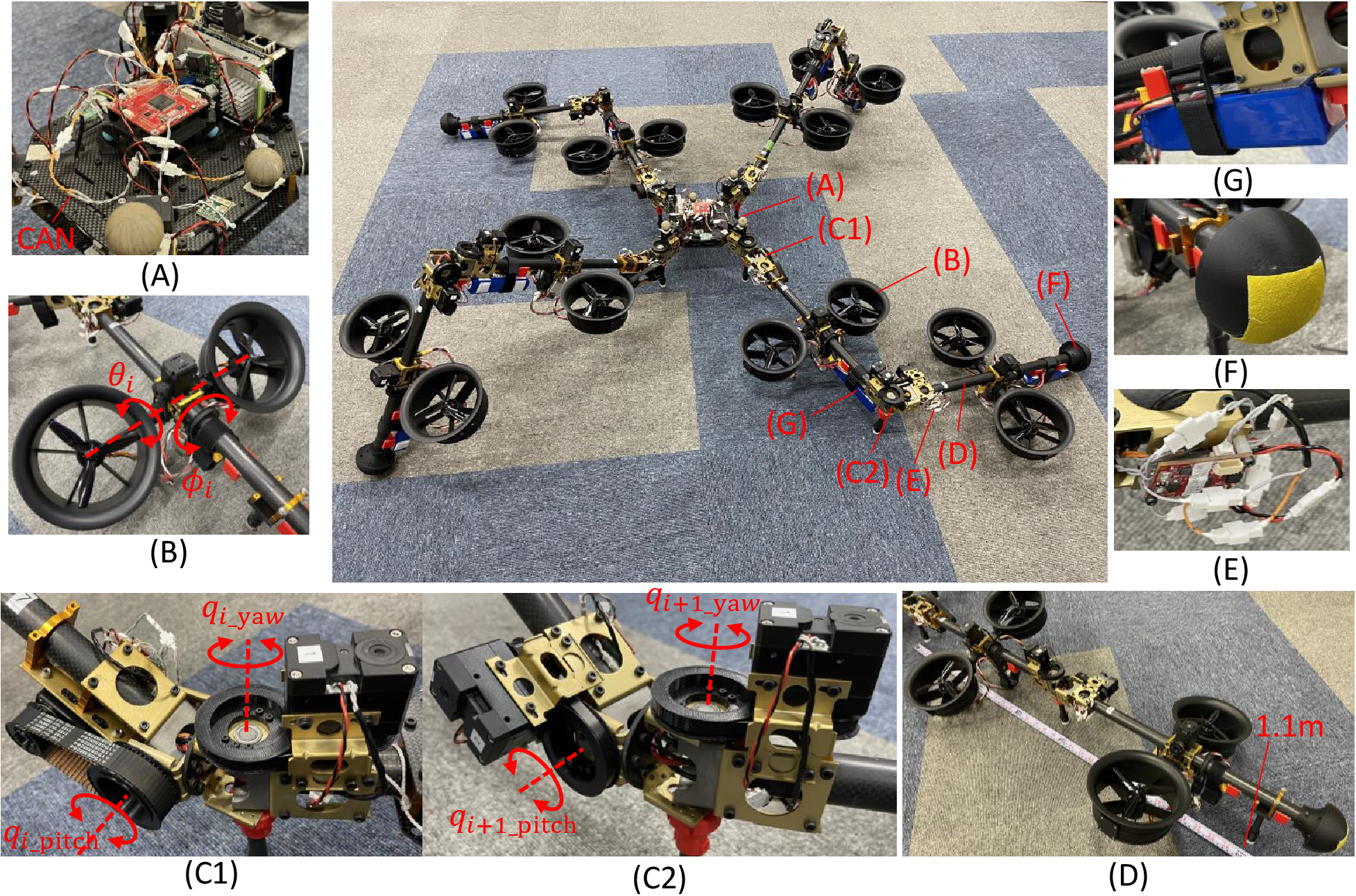}
    \vspace{-5mm}
    \caption{{\bf Prototype of SPIDAR}: {\bf (A)} center torso that employed an original red MCU called {\it Spinal} and a high level processor (Nvidia Jetson TX2); {\bf (B)} spherically vectorable dual-rotor module; {\bf (C1)(C2)} two-DoF joint module for the ``hip'' and the ``knee'', respectively; {\bf (D)} single leg (limb) that had the maximum length of \SI{1.1}{m}; {\bf (E)} small relay board called ``Neuron'' for each link unit that was connected with ``Spinal'' via CAN; {\bf (F)} hemisphere foot with anti-slip tape; {\bf (G)} distributed battery attached at each link unit.}
    \label{figure:platfrom}
  \end{center}
\end{figure}

In this work, we developed a prototype of SIDAR as shown in \figref{figure:platfrom}, and the basic specification is summarized in \tabref{table:specification}. Given the lightweight design, we employed CFRP material for link rod where cables can pass through. For the joint module, we used the Aluminum sheet to connect links, whereas the joint servo was Dynamixel XH430-V350R of which the torque was enhanced by pulley made from PLA. The range of joint angle was $\left[-90^{\circ} \ 90^{\circ}\right]$.
For the vectorable rotor module, a pair of counter-rotating plastic propellers were enclosed by ducts with the aim of safety and increase of thrust, whereas  Dynamixel XL430-W250T was used for the rotor vectoring.
Batteries are distributed in each link unit in parallel as shown in \figref{figure:platfrom}(G) which can provide a flight duration up to \SI{9}{min} and a longer walk duration up to \SI{20}{min}. A hemisphere foot with anti-slip tape was equipped to ensure the stable point contact during the terrestrial locomotion.

On the center torso as shown in \figref{figure:platfrom}(A), NVIDIA Jetson TX2 and an original MCU board called {\it Spinal} were deployed to perform the realtime control framework as presented in \figref{figure:control}. For each link unit, there was a distributed MCU board called {\it Neuron} that served as relay node between {\it Spinal} and each actuator. {\it Neuron}s and {\it Spinal} were connected by CAN cable. The detail of the onboard communication can be found in \cite{aerial-robot:DRAGON-RAL2018}. Besides, an external motion capture system was applied in our experiment to obtain the state of the baselink (i.e., $\bm{r}_b$, $\dot{\bm r}_b$, $R_b$, and ${\bm \omega}_b$), which were used to calculate the state of centroidal motion based on forward-kinematics.

\begin{table}[t]
  \begin{center}
    \caption{Prototype Specifications}
    \begin{tabular}{c|ccc|c}
      \multicolumn{2}{c}{1. Main Feature} &  & \multicolumn{2}{c}{3. Vectorable Rotor} \\
      Attribute & Value  && Attribute & Value \\ \cline{1-2} \cline{4-5}
      total mass & \SI{15.2}{kg} && rotor KV &  1550 \\ \cline{1-2} \cline{4-5}
      max size (dia.) & \SI{2.7}{m} && propeller diameter & \SI{5}{inch} \\ \cline{1-2} \cline{4-5}
      max flight time &  \SI{9}{min} && max thrust ($\bar{\lambda}$) & \SI{42}{N} \\ \cline{1-2} \cline{4-5}
      max walk time &  \SI{20}{min} && pulley ratio & 1:1.5  \\ \cline{4-5}
      \multicolumn{2}{c}{} && max vectoring torque & \SI{1.5}{Nm} \\ \cline{4-5}
      \multicolumn{2}{c}{2. Link and Joint} && max vectoring speed & \SI{4.2}{rad/s} \\  \cline{4-5}
      Attribute & Value && \multicolumn{2}{c}{} \\ \cline{1-2} \cline{1-2}
      link length & \SI{0.54}{m} && \multicolumn{2}{c}{4. Lipo Battery} \\ \cline{1-2}
      joint pulley ratio & 1:2 && Attribute & Value \\ \cline{1-2} \cline{4-5}
      max joint speed & \SI{0.34}{rad/s} && capacity & 6S 3Ah \\ \cline{1-2} \cline{4-5}
      max torque ($\bar{\tau}_q$)  & \SI{6.5}{Nm}  && amount &  8 \\ \cline{1-2} \cline{4-5}
    \end{tabular}
    \label{table:specification}
  \end{center}
\end{table}

\subsection{Basic Experimental Evaluation}


\subsubsection{Aerial transformation}

\begin{figure}[b]
  \begin{center}
    \includegraphics[width=1.0\columnwidth]{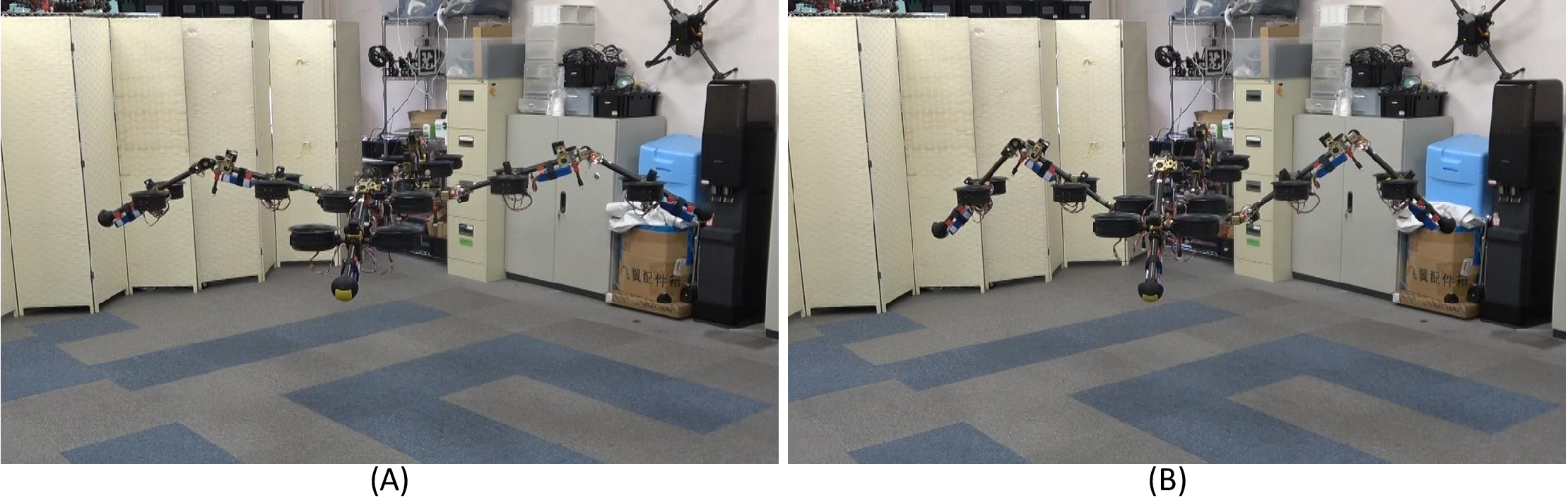}
    \vspace{-5mm}
    \caption{{\bf Stable joint motion in midair}: {\bf (A)} extended pose that has diameter of \SI{2.6}{m}; {\bf (B)} standing pose, implying the feasibility to takeoff directly from the terrestrial mode.}
    \label{figure:flight}
  \end{center}
  \vspace{-5mm}
\end{figure}
\begin{figure}[!b]
  \begin{center}
    \includegraphics[width=1.0\columnwidth]{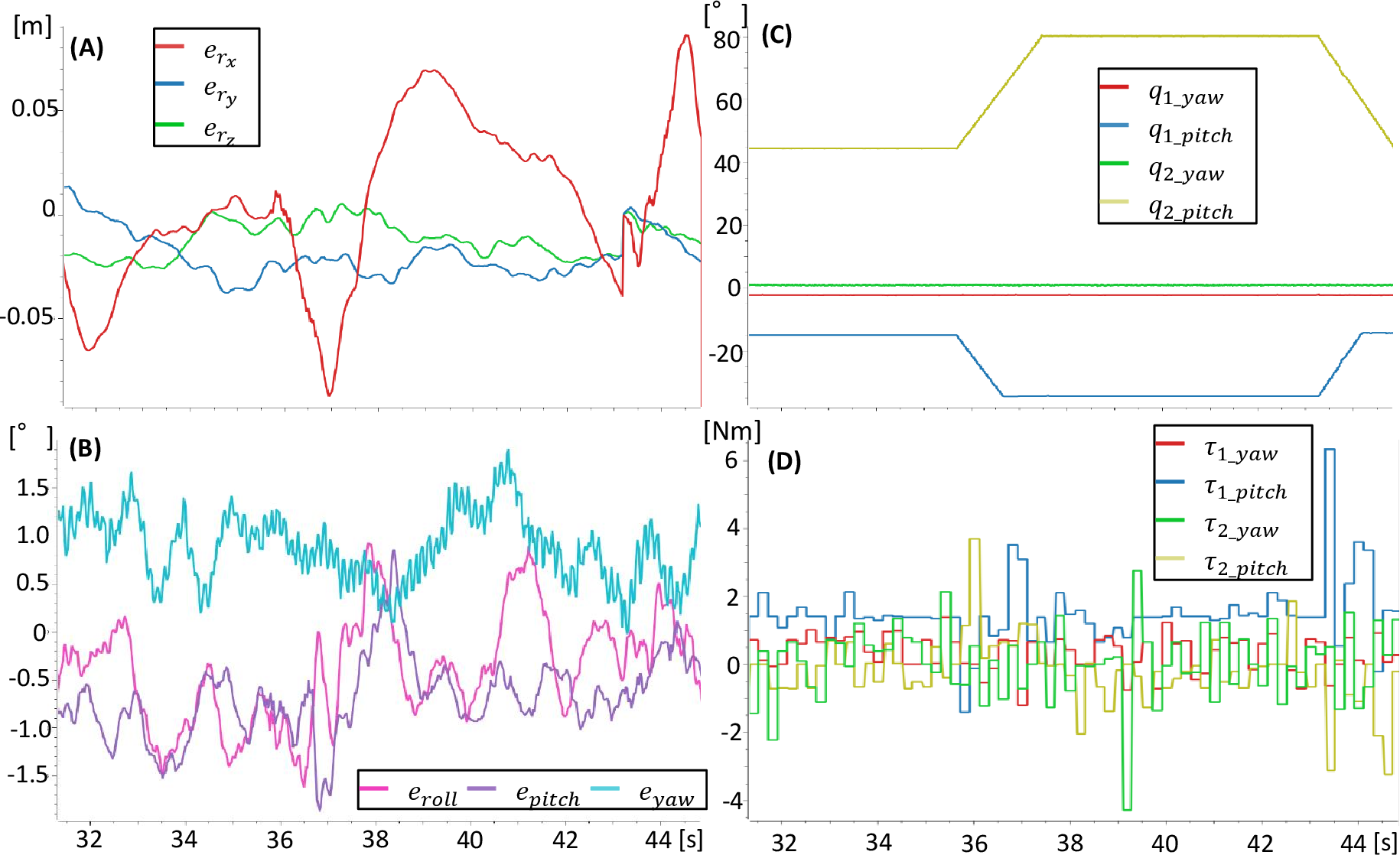}
    \vspace{-3mm}
    \caption{{\bf Plots related to \figref{figure:flight} }: {\bf (A)} positional errors  of $\{CoG\}$; {\bf (B)} rotational errors of $\{CoG\}$ described in XYZ Euler angles; {\bf (C)} joint trajectories for leg1 ($q_{1\_{\mathrm{yaw}}}$, $q_{1\_{\mathrm{pitch}}}$, $q_{2\_{\mathrm{yaw}}}$, and $q_{2\_{\mathrm{pitch}}}$), other legs followed the same joint trajectories; {\bf (D)} torques for those joints.}
    \label{figure:flight_plot}
  \end{center}
\end{figure}

A unique feature of SPIDAR is the aerial maneuvering with joint motion (i.e., aerial transformation). To validate the stability during flight, simple transformation as shown in \figref{figure:flight} was performed. All limbs changed their joints with the same trajectories as plotted in \figref{figure:flight_plot}(C).
For the control gains in \equref{eq:pid_pos} and \equref{eq:pid_rot}, we set $K_{f, p}$, $K_{f, i}$, $K_{f, d}$, $K_{\tau, p}$, $K_{\tau, i}$, and $K_{\tau, d}$ as $D(3.6, 3.6, 2.8)$, $D(0.03, 0.03, 1.2)$, $D(4, 4, 2.8)$, $D(15, 15, 10$), $D(0.3, 0.3, 0.1)$, and $D(5,5,5)$, where $D(*,*,*) \in \mathcal{R}^{3\times3}$ is a diagonal matrix. For the optimization problem of \equref{eq:rough_allocation_cost}, we omitted the second term (i.e., $w_2 = 0$) to put a priority on the minimization of the thrust force. 
\figref{figure:flight_plot}(A) and (B) plotted the positional and rotational errors during the flight and transformation, and the RMS of those errors were [0.014, 0.023, 0.038] \si{m} and [0.81, 0.69, 0.92]$^{\circ}$. The altitude error $e_{r_z}$ indicated a relatively large deviation during the joint motion, which was caused by the violation of the quasi-static assumption. Nevertheless, this deviation rapidly decreased once the joint motion finished. \figref{figure:flight_plot}(C) and (D) showed that all joint were well controlled by the PD control as presented in \equref{eq:joint_pd_control}. Eventually, these results demonstrated the stability of both the baselink pose and the joint motion in aerial locomotion.


\subsubsection{Leg lifting}

\begin{figure}[t]
  \begin{center}
    \includegraphics[width=1.0\columnwidth]{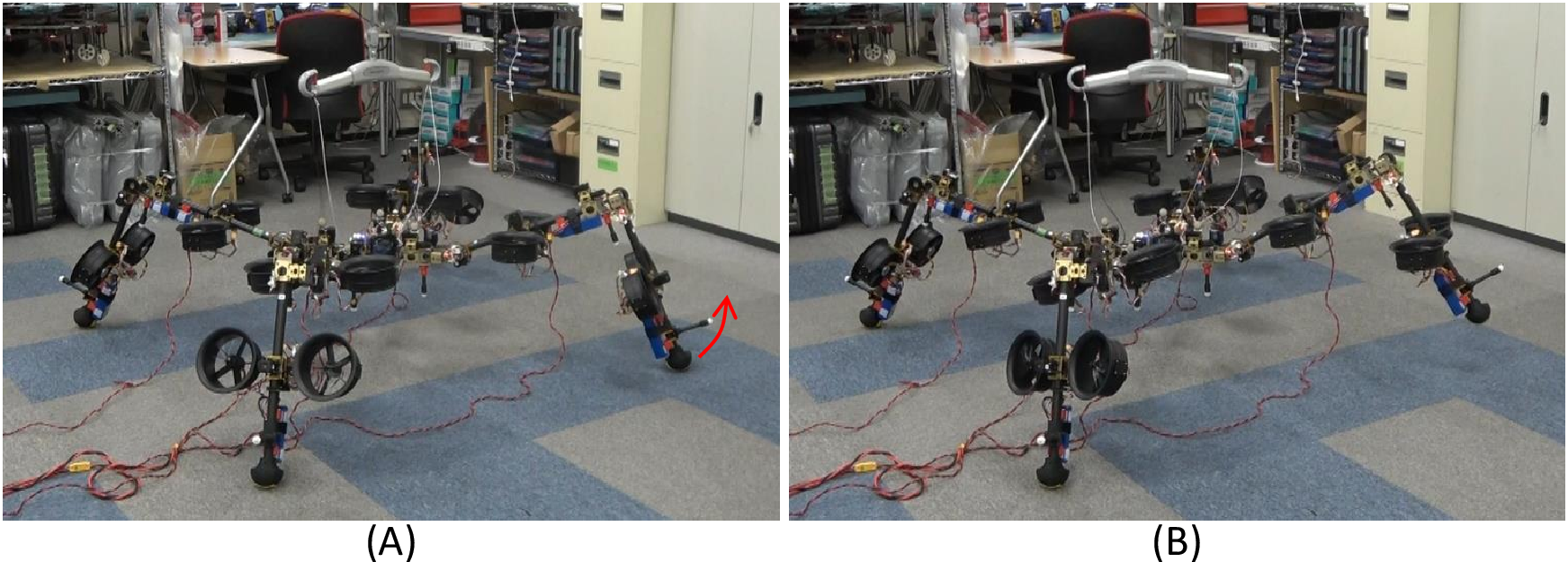}
    \vspace{-5mm}
    \caption{{\bf lifting a single leg from standing mode}: {\bf (A)} standing mode where all feet have contact with the ground; {\bf (B)} lifting single leg and keeping the raised pose with the assistance of rotor thrust.}
    \label{figure:raise_test}
  \end{center}
\end{figure}

\begin{figure}[t]
  \begin{center}
    \includegraphics[width=1.0\columnwidth]{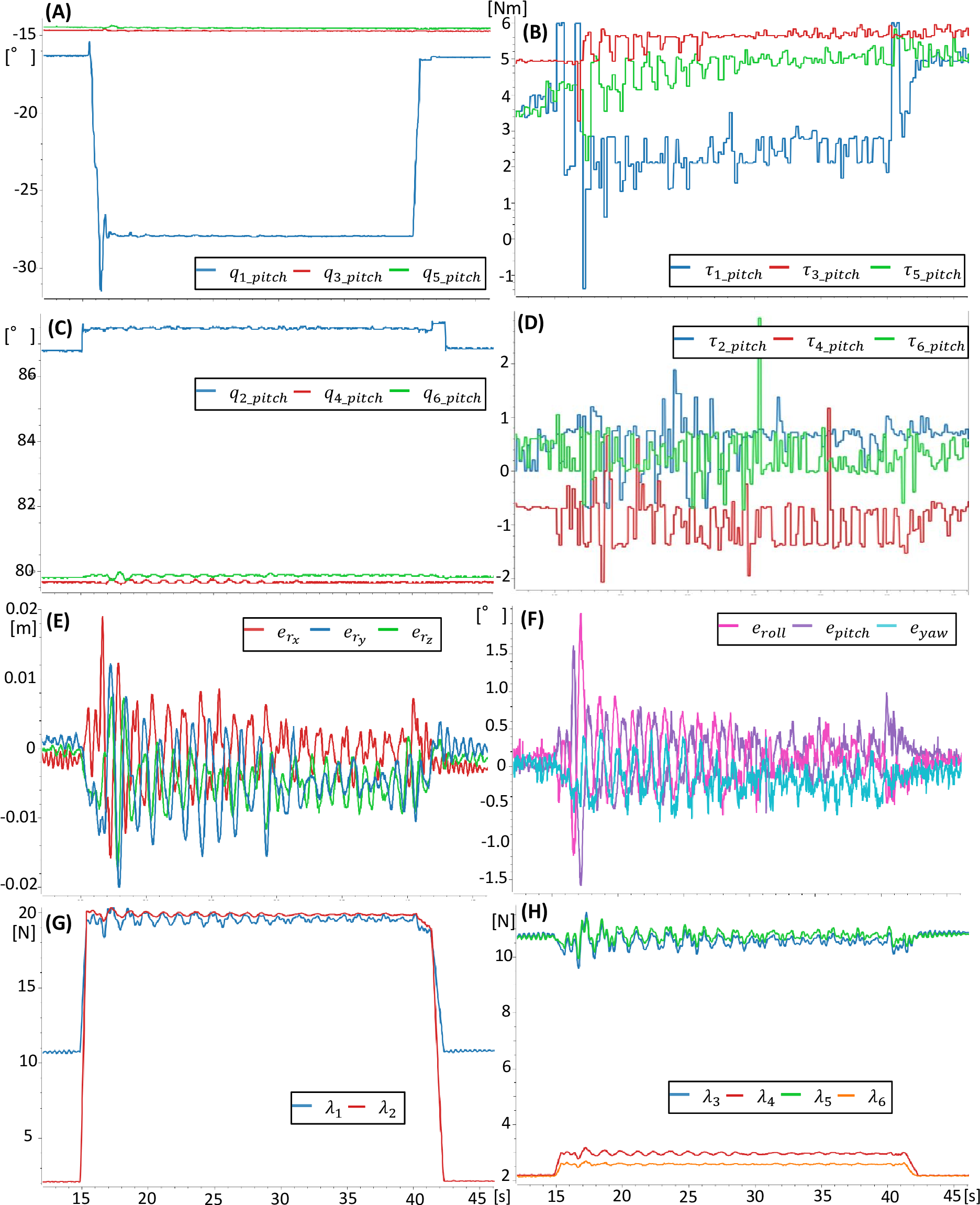}
    \vspace{-5mm}
    \caption{{\bf Plots related to \figref{figure:raise_test}}: {\bf (A)} trajectories for hip pitch joints. $q_{7\_{\mathrm{pitch}}}$ was omitted due to the symmetric pose of leg4 related to leg2; {\bf (B)} torques of joints in (A); {\bf (C)} trajectories for knee pitch joints; {\bf (D)} torques of joints in (C); {\bf (E)} positional errors of baselink ; {\bf (F)} rotational errors of baselink; {\bf (G)} thrust forces in leg1; {\bf (H)} thrust forces in other legs.}
    \label{figure:raise_plot}
  \end{center}
  \vspace{-5mm}
\end{figure}

The key to achieve walking by legged robot is the stability while lifting the leg.
Then, we evaluated the proposed control method by performing a long-term single leg lifting as shown in \figref{figure:raise_test}. The cost weights in \equref{eq:rough_allocation_cost} were set as $w_1 = 1, w_2 = 1$, and the bound for joint torque $\bar{\tau}_{q}$ was decreased to $\SI{1.5}{Nm}$ to ensure sufficient margin for joint control. Besides, the gain $k_{b}$ in \equref{eq:pid_torso_altitude} was set as 25.
Leg1 was lifted by changing $q_{1\_{\mathrm{pitch}}}$ from \SI{-16}{^{\circ}} to \SI{-28}{^{\circ}}, and the lifting motion lasted around \SI{30}{s} as shown in \figref{figure:raise_plot}(A). Other joints were kept constant in the whole motion as shown in \figref{figure:raise_plot}(A) and (C), and their torques were within the bounds as depicted in \figref{figure:raise_plot}(B) and (D). These results demonstrated the stability of joint motion against the influence of thrust force. Besides the stability of the baselink pose can be confirmed in \figref{figure:raise_plot}(E) and (F), where both the positional and rotational errors converged to the sufficiently small value (i.e., \SI{0.01}{m} and \SI{0.5}{^{\circ}}). \figref{figure:raise_plot}(G) showed the large increase of the thrust forces in the lifting leg, whereas \figref{figure:raise_plot}(H) showed small changes in other standing legs. In addition, these plots also confirmed the stable transition between standing mode and leg lifting mode. In particular, the shift back to the standing model around \SI{40}{s} demonstrated the smooth touchdown, which indicates the promising terrestrial locomotion.




\subsection{Seamless Terrestrial/Aerial Hybrid Locomotion}
\label{subsec:seamless_motion}

\begin{figure}[!b]
  \begin{center}
    \includegraphics[width=1.0\columnwidth]{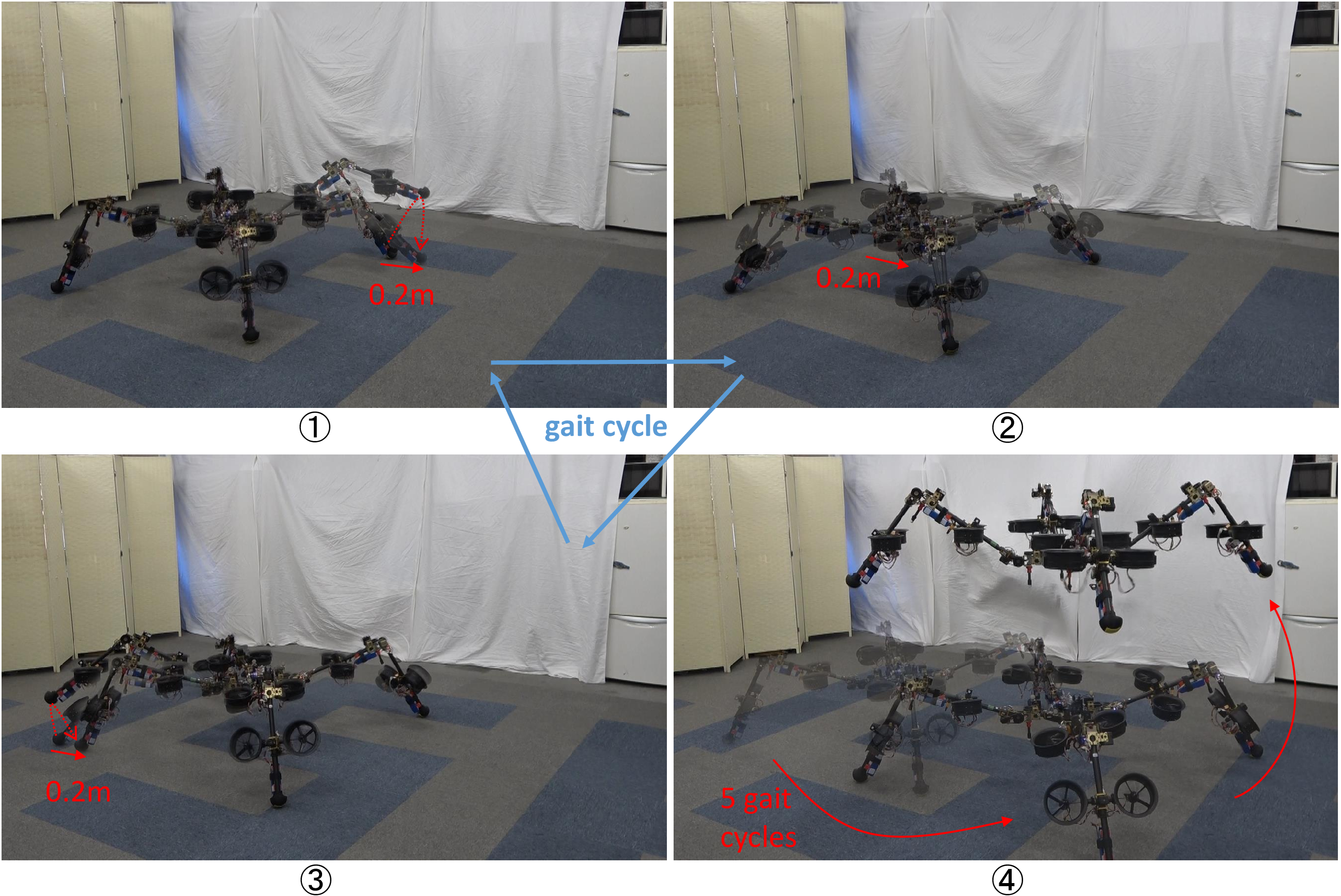}
    \vspace{-5mm}
    \caption{{\bf Seamless Terrestrial/Aerial Hybrid Locomotion}: \textcircled{\scriptsize 1} $\sim$ \textcircled{\scriptsize 3} shows the representative phases (moving the front-left leg, the torso, and rear-left leg) in one creeping gait cycle. After five gait cycles, robot switched to the aerial locomotion directly from the terrestrial pose as shown in \textcircled{\scriptsize 4}.}
    \label{figure:seamless_motion}
  \end{center}
\end{figure}

\begin{figure}[t]
  \begin{center}
    \includegraphics[width=1.0\columnwidth]{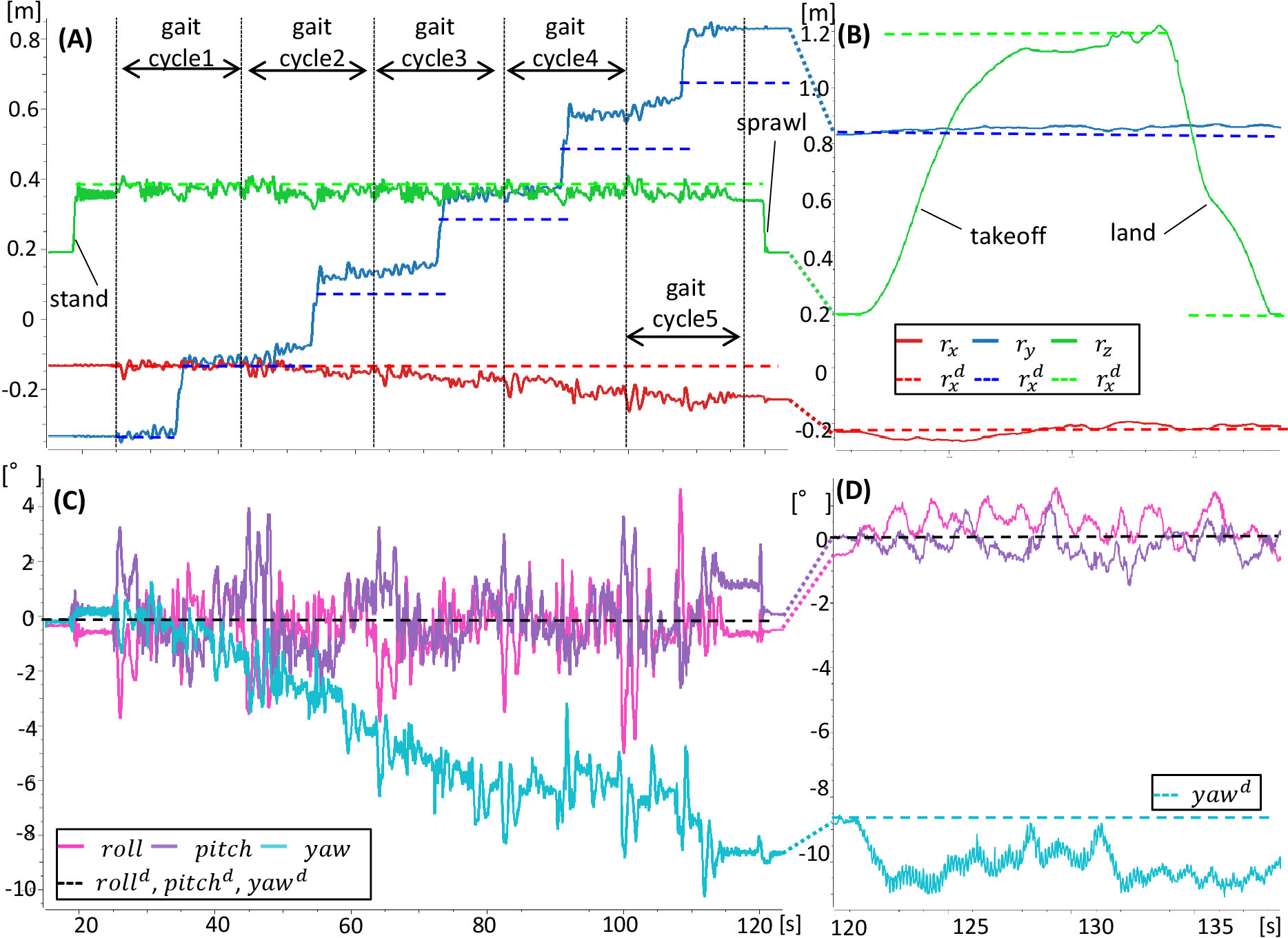}
    \vspace{-5mm}
    \caption{{\bf Plots related \figref{figure:seamless_motion}}.{\bf (A)/(B)} trajectories of baselink position during the terrestrial locomotion and the aerial locomotion, respectively; {\bf (C)/(D)} trajectories of baselink orientation during the terrestrial locomotion and the aerial locomotion, respectively.}
    \label{figure:seamless_motion_plot}
  \end{center}
\end{figure}

We further evaluated the feasibility of seamless locomotion transition as shown in \figref{figure:seamless_motion}. \figref{figure:seamless_motion_plot}(A) and (C) demonstrated the baselink pose trajectory during walking with five gait cycles. We observed that the translational drift along the walking direction ($x$ axis) and the orthogonal direction ($y$ axis) finally grew to \SI{0.18}{m} and \SI{0.10}{m}, whereas the rotational drift along the yaw axis also increased to \SI{9}{^{\circ}}. These drifts can be attributed to the feed-froward gait planing where the target baselink pose was updated based on the last target values but not the actual values. Nevertheless, these drifts can be considered relatively small compared to the total displacement, and are possible to be suppressed by adding a feed-back loop in planning as a future work. Furthermore, the deviations regarding the $z$, roll, and pitch axes were sufficiently small, which demonstrated the efficiency of the proposed control method presented in \secref{control}.

As shown in \figref{figure:seamless_motion_plot}(B) and (D), the transition to the aerial locomotion was smooth and stable, and the stability in midair was also confirmed, Thus, these results demonstrated the feasibility of the mechanical design, modeling and control methods for the terrestrial/aerial hybrid quadruped platform.


\section{Conclusion}
\label{sec:conclusion}

In this paper, we presented the achievement of the terrestrial/aerial hybrid locomotion by the quadruped robot SPIDAR that were equipped with the vectorable rotors distributed in all links. We first proposed the mechanical design for this unique quadruped platform, and then developed the modeling and control methods to enable static walking and transformable flight. The feasibility of the above methods were verified by the experiment of seamless terrestrial/aerial hybrid locomotion with the prototype of SPIDAR.

A crucial issue remained in this work is the oscillation and deviation of the baselink pose and joint angles during walking. To improve the stability, the rotor thrust should be directly used in the joint position control to replace the current simple PD control.
Furthermore, the gait planning should be also robust against the drift by adding a feed-back loop as discussed in \subsecref{seamless_motion}.
Last but not least, the dynamic walking and the aerial manipulation will be investigated to enhance the versatility of this robot in both maneuvering and manipulation.



\bibliographystyle{junsrt}
\bibliography{main}

\end{document}